\documentclass[conference]{IEEEtran}
\usepackage{cite}

\ifCLASSINFOpdf
   \usepackage[pdftex]{graphicx}
\else
\fi

\usepackage{stfloats}

\begin{document}
%
\title{Games and Big Data: A Scalable \\ Multi-Dimensional Churn Prediction 
Model}



\author{\IEEEauthorblockN{Paul Bertens, Anna Guitart and \'{A}frica 
Peri\'a\~{n}ez}
\IEEEauthorblockA{
Game Data Science Department\\
Silicon Studio\\
1-21-3 Ebisu Shibuya-ku, Tokyo, Japan\\
\{paul.bertens, anna.guitart, africa.perianez\}@siliconstudio.co.jp}
}


%
%


\maketitle

\begin{abstract}
The emergence of mobile games has caused a paradigm shift in the video-game 
industry. Game developers now have 
at their disposal a plethora of information on their players, and thus can take advantage of 
reliable models that can accurately predict 
player behavior and scale to huge datasets. Churn prediction, a challenge common 
to a variety of sectors, is particularly 
relevant for the mobile game industry, as player retention is crucial for the 
successful monetization of a game. In this 
article, we present an approach to predicting game abandon based on survival 
ensembles. Our method provides accurate 
predictions on both the level at which each player will leave the game and their 
accumulated playtime until that moment. Further, 
it is robust to different data distributions and applicable to a wide range 
of response variables, while also allowing for efficient parallelization 
of the algorithm.  This makes our model well suited to perform 
real-time analyses of churners, even for games with millions of daily active 
users. 
\end{abstract}


\section{Introduction}
\label{intro}
In the last few years, the video-game industry has been shaken by the appearance 
of mobile games. Currently, 
both traditional console games and mobile games are always online and allow game 
developers to record every 
action of the players. Such a unique source of information opens the door to 
achieving a comprehensive analysis 
of player behavior and a full understanding of player needs on quantitative 
grounds.

Preventing user abandon is a challenge faced by many industries  and especially 
relevant for the video-game sector. Indeed, 
acquisition campaigns to obtain new players are expensive, while retaining 
existing users is more cost-effective.
Identifying churners beforehand allows game owners to perform personalized 
promotion campaigns to retain the most valuable players 
and efficiently increase monetization. Even though there have been some works on 
modeling churn in the field of mobile 
games \cite{runge2014, hadiji,rothenbuehler2015hidden}, they generally use 
techniques that either make binary predictions, rely 
on models that are not readily applicable to different data distributions or are 
not able to capture the temporal dynamics 
intrinsic to churn. They also present some drawbacks regarding scalability.

In this paper we discuss churn prediction beyond the classical binary approach. 
Previous works\cite{perianez2016churn} have shown how to predict player abandon 
in terms of days, i.e.\ \emph{time-to-event}, 
using survival analysis embedded into ensemble modeling. The present study 
introduces, for the first time in the mobile-game sector, 
a model that accurately predicts the level at which a player is expected to 
leave the game and their hours of playtime until that moment. 
Our methodology allows for a comprehensive solution to the churn prediction 
challenge from several perspectives and dimensions, helping to fully understand 
and anticipate player attrition.

\begin{figure}[ht!]
  \centering
  \includegraphics[width=0.24\textwidth]{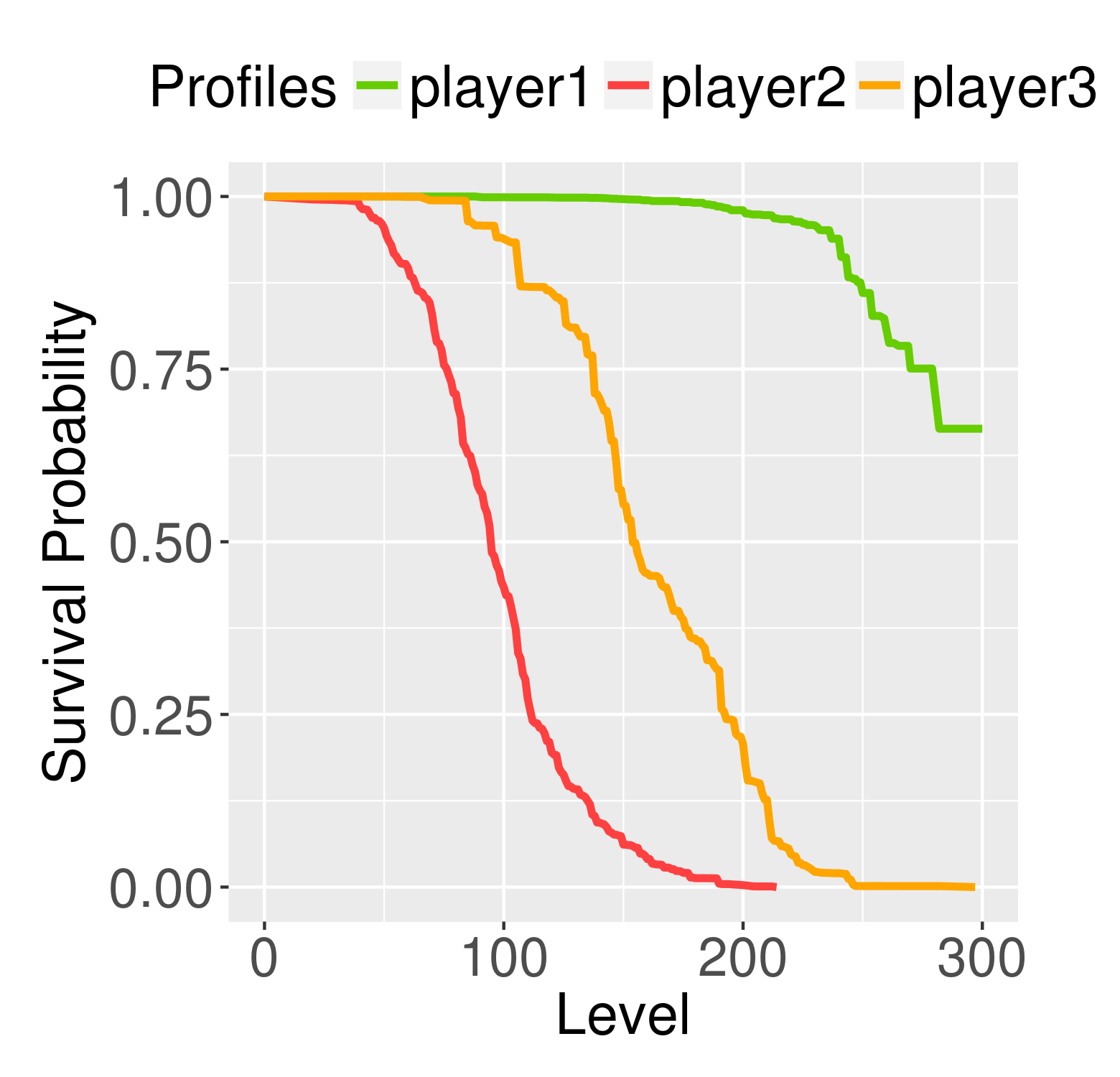}
   \includegraphics[width=0.24\textwidth]{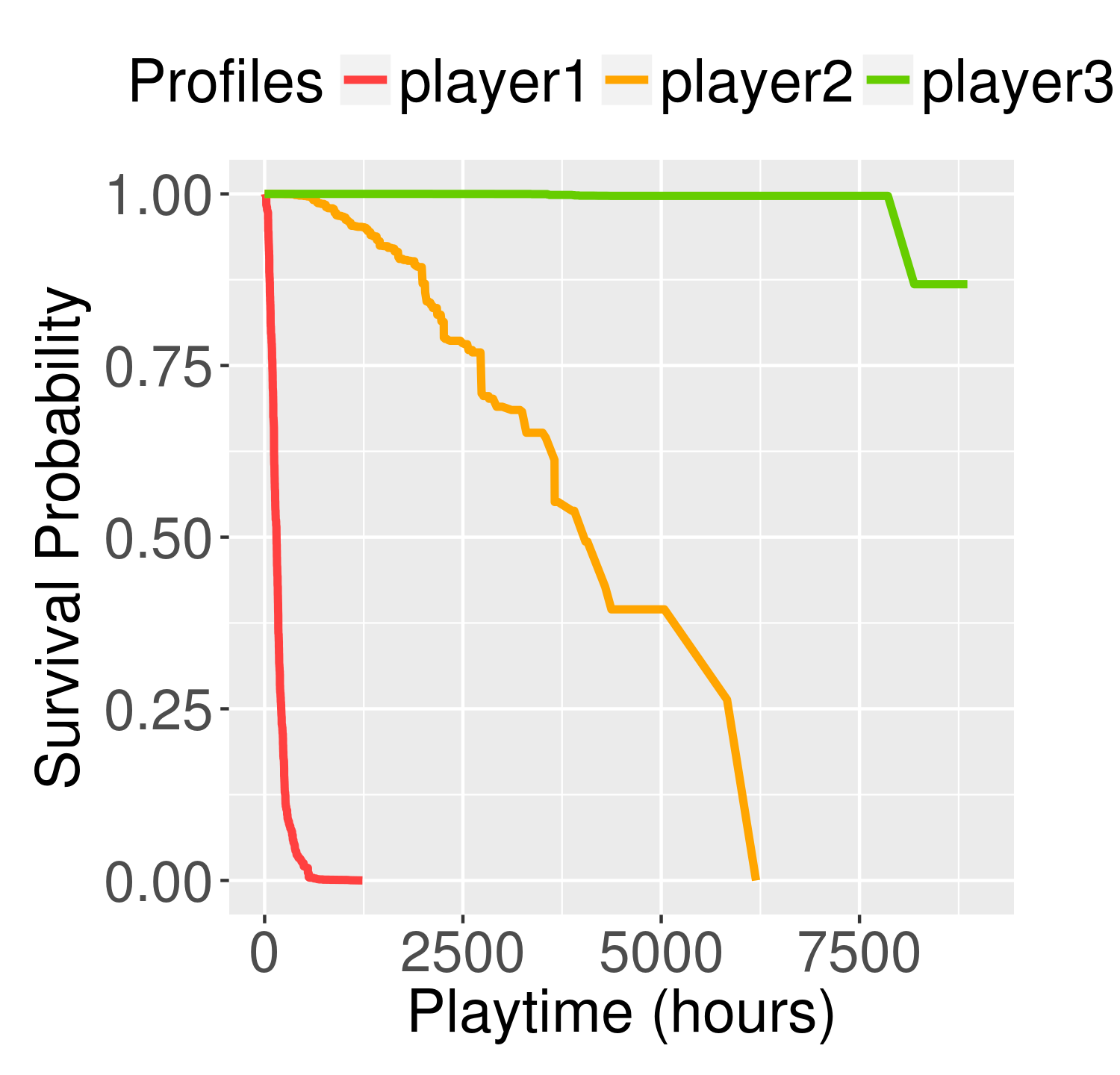} 
\caption{Predicted survival probability by level and playtime for three players. The 
first player is expected to churn at approximately level 100 and play about 500 hours (red), the second around level 200 and will play 
5000 hours (yellow), and a loyal player who is not predicted to leave (green).}
\label{level-users}
\end{figure}

%

%

\section{Method}
\subsection{Model}
The method presented here is an extension of previous work on churn prediction 
in mobile social games \cite{perianez2016churn} using \textit{conditional inference survival ensembles} 
\cite{Hothorn06unbiasedrecursive}. Based on survival analysis \cite{clark}, the model is 
capable of performing accurate predictions even when the response variable is 
censored. 

\begin{figure*}[ht!]
  \centering
  \includegraphics[width=0.49\textwidth]{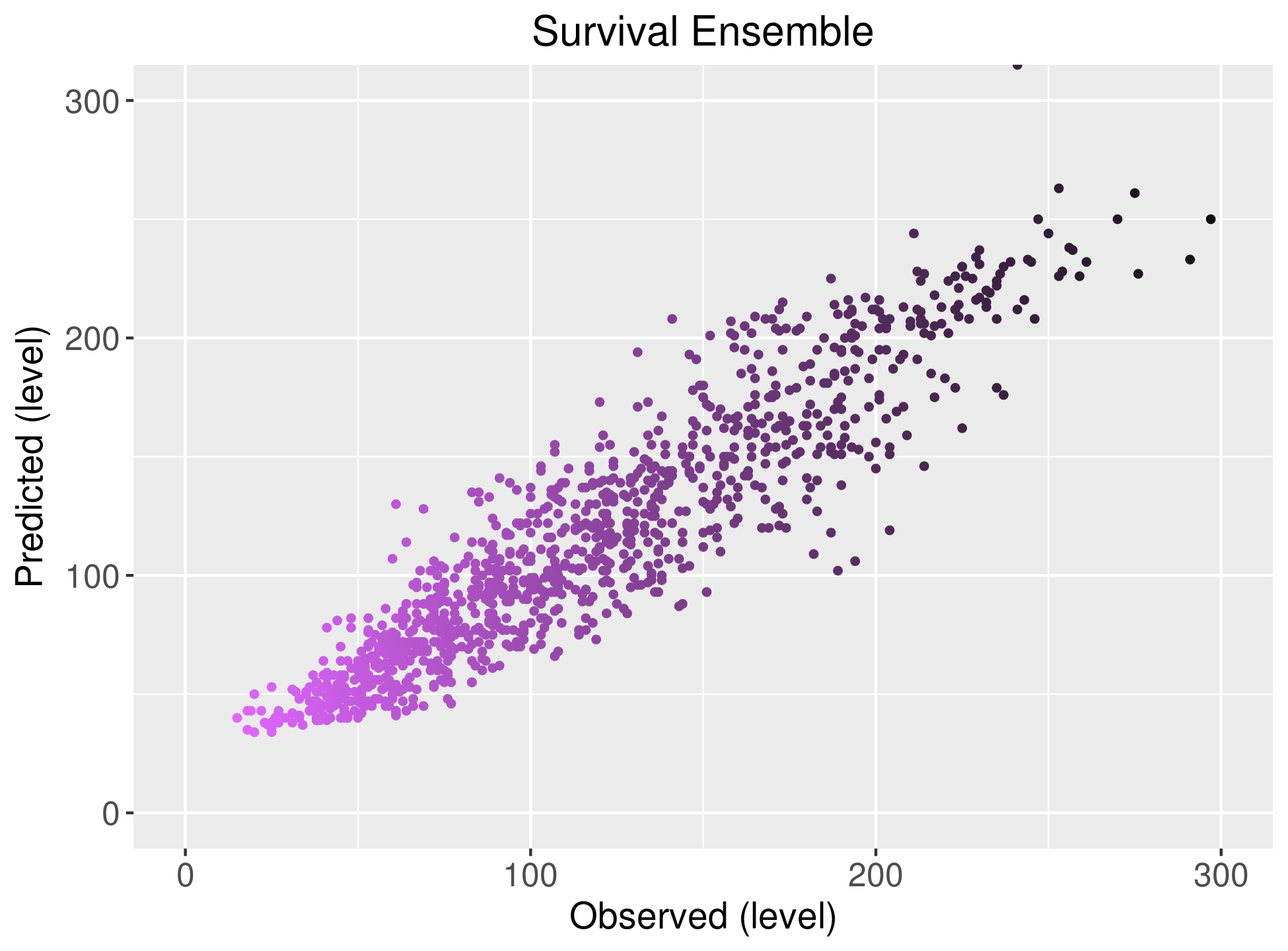}
   \includegraphics[width=0.49\textwidth]{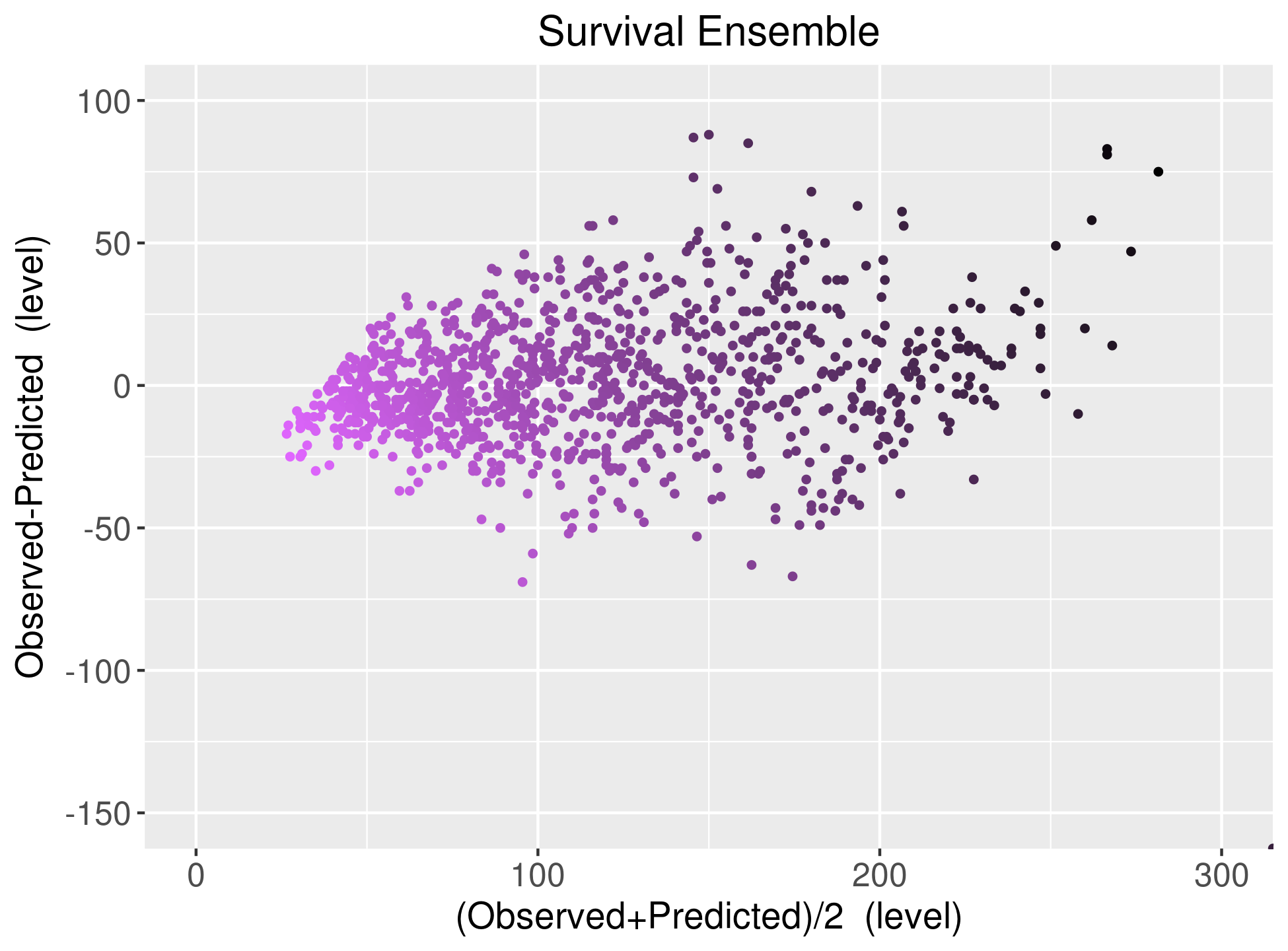} \\
     \includegraphics[width=0.49\textwidth]{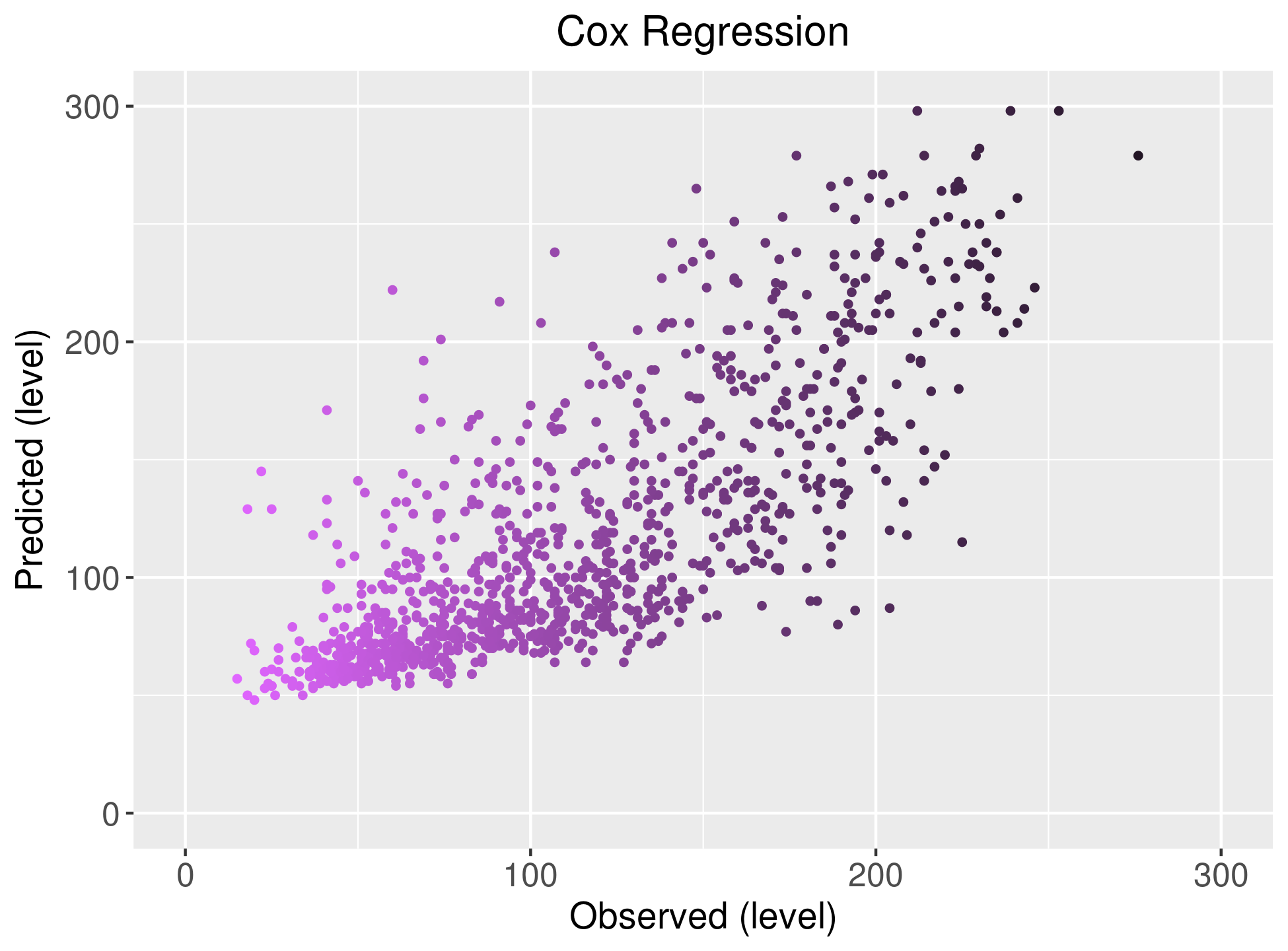}
   \includegraphics[width=0.49\textwidth]{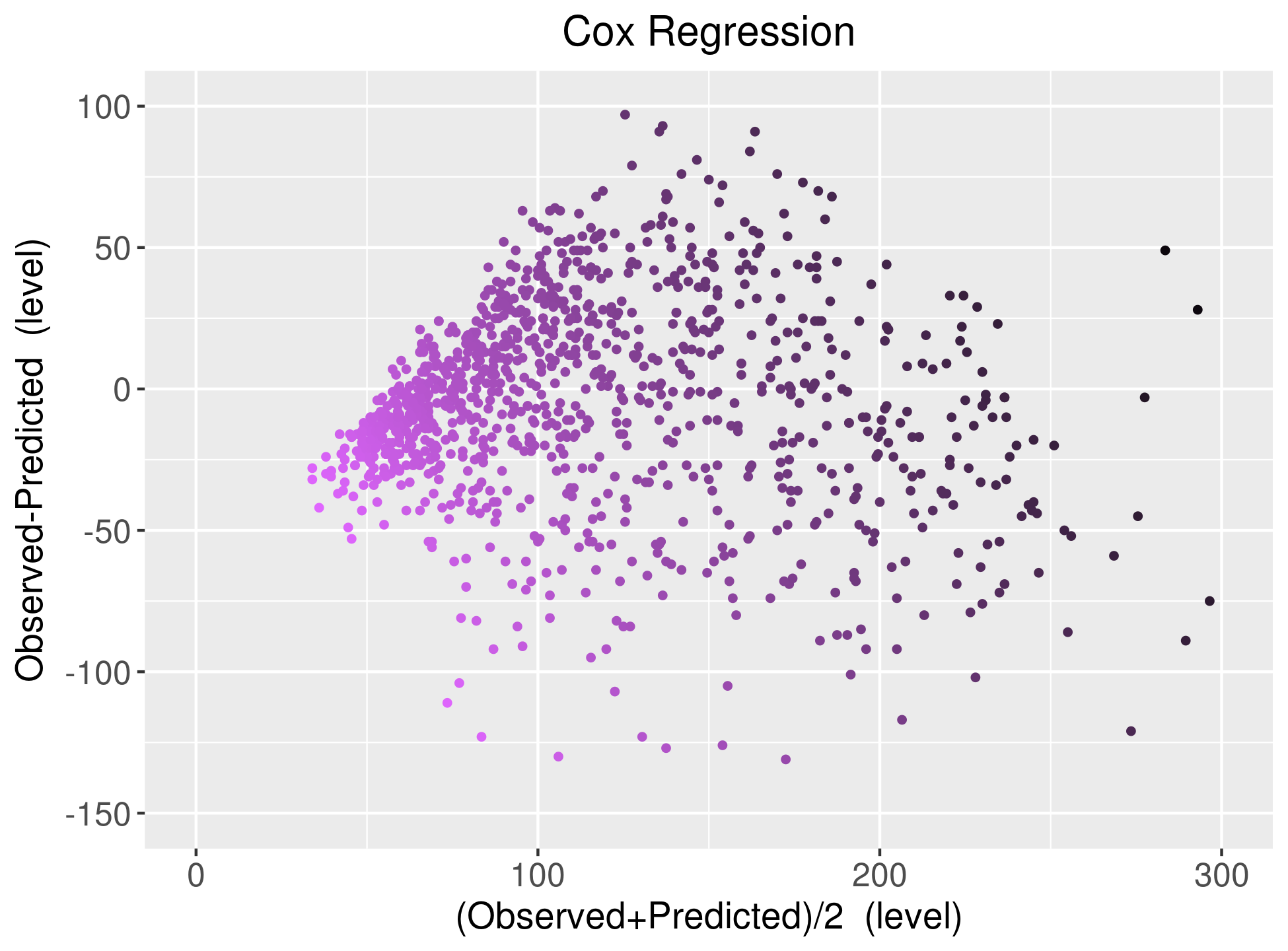} \\
\caption{ Predicted median survival level vs.\ observed level (left) and 
relative deviation (right) for churned players, using the survival ensemble and 
Cox regression models}
\label{level}
\end{figure*}

A survival ensemble is an ensemble of survival trees. Every tree calculates weighted Kaplan--Meier estimates to distinguish the different survival characteristics of every sample in the tree nodes.
Linear rank statistics are used as splitting criterion of the nodes, in order to maximize the survival difference among the daughter nodes. Because the 
partitions of every tree are computed in two steps, the \textit{conditional inference survival ensembles} \cite{Hothorn06unbiasedrecursive} are not
biased towards predictors with many splits and are more robust to overfitting. First, the optimal split variable is selected 
based on the relationship between the covariates and response and then the optimal split point is obtained by comparing two sample linear statistics for all possible partitions of the split covariate.

We have implemented a parallelized version that is more practical in a 
production setting and can also make predictions on other response variables, 
including level and playtime. 
The approach taken here is to parallelize computations using multiple cores on a 
single machine. Each core trains a subset of trees from the total ensemble, and 
at the end of training all trees are merged to obtain the final model. This 
method can be easily extended to run on multiple machines, where each machine 
takes a subset of the total ensemble and stores the trained partial models on a 
shared disk, finally merging them back into a single model. 

A similar parallelization can be used to obtain accurate predictions on individual players: each 
core focuses on just a partial subset of players, and full-survival probability 
curves for each user are efficiently computed across multiple cores.

\begin{figure*}[ht!]
  \centering
  \includegraphics[width=0.49\textwidth]{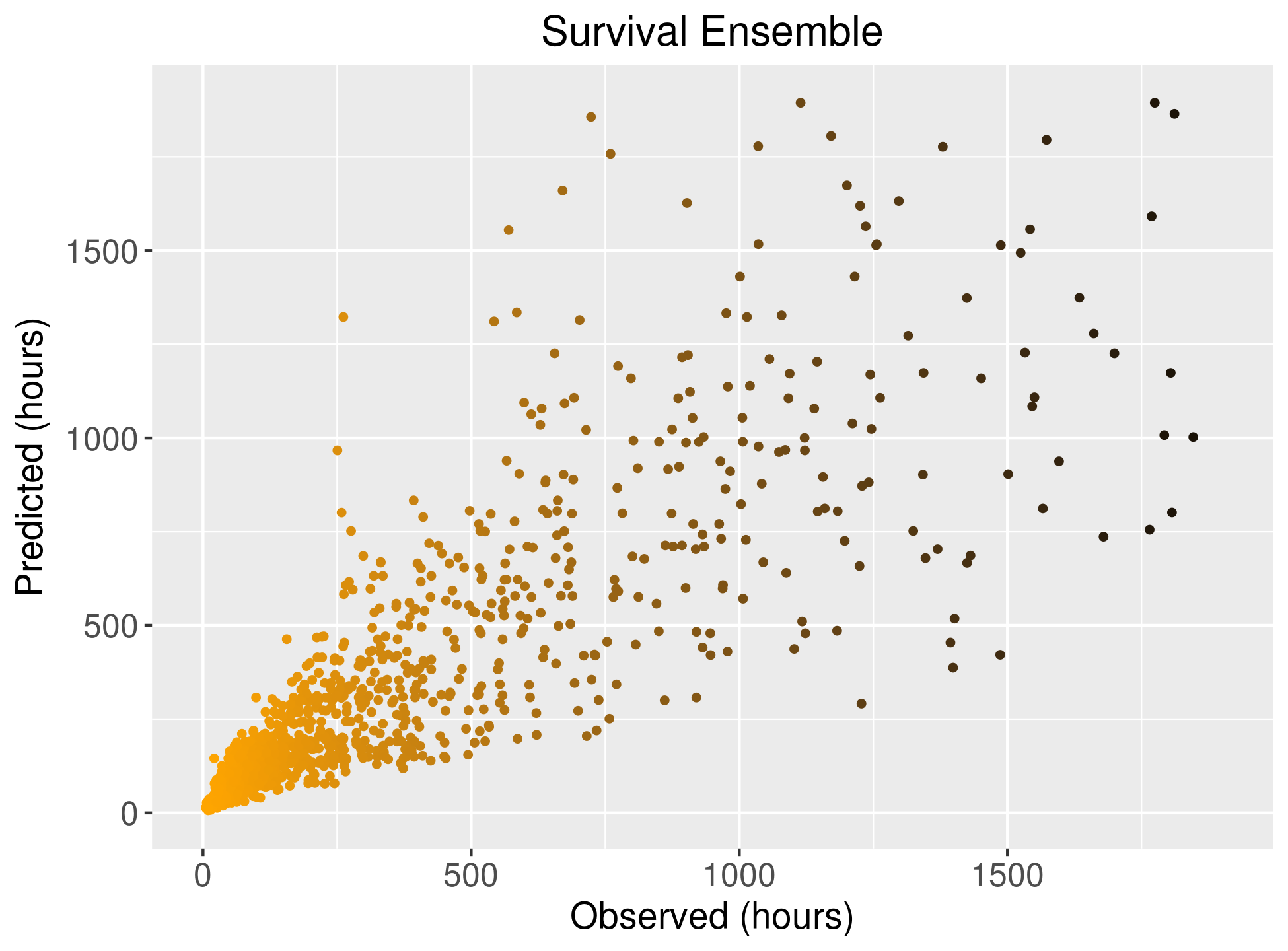}
   \includegraphics[width=0.49\textwidth]{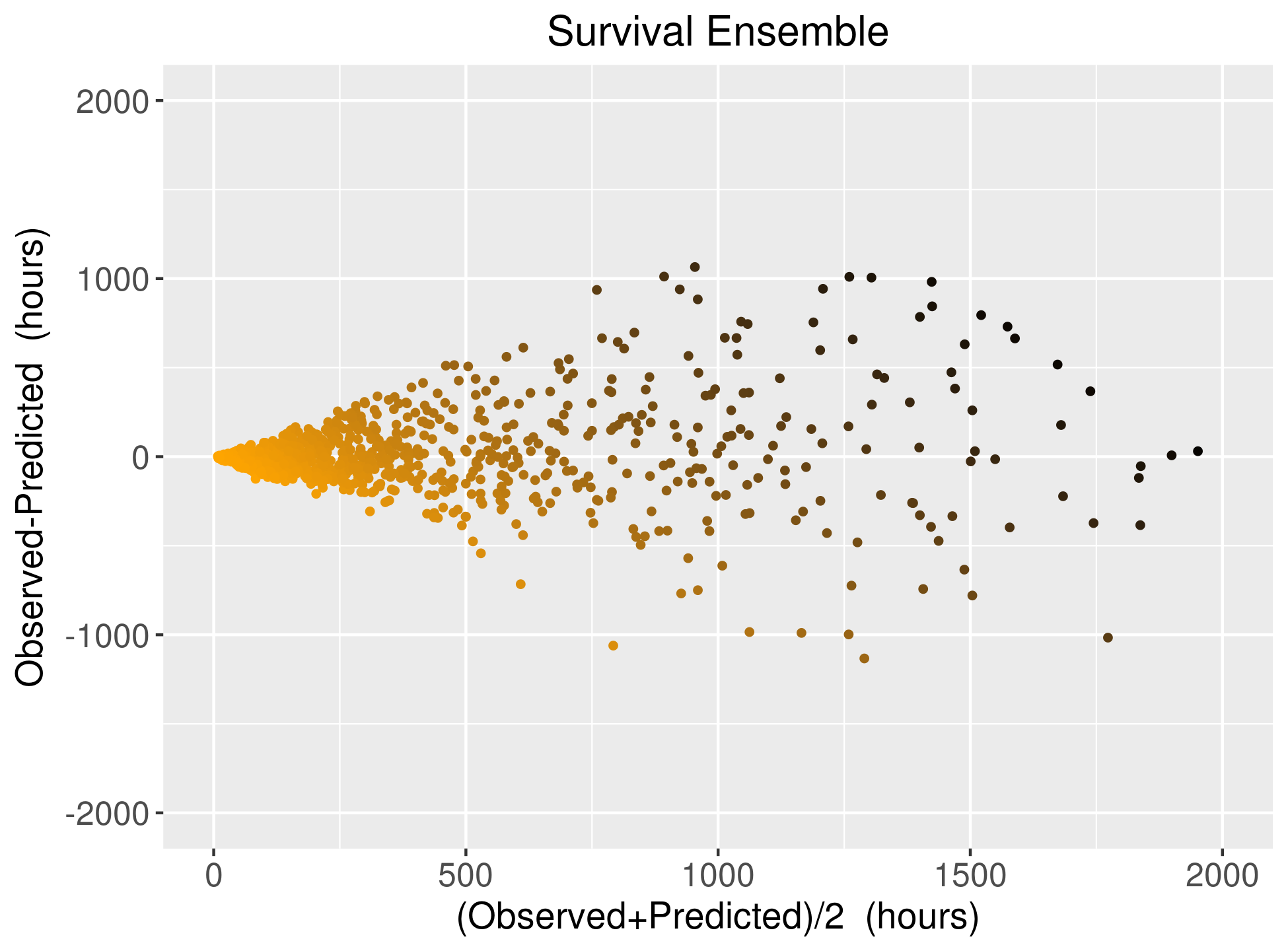} \\
     \includegraphics[width=0.49\textwidth]{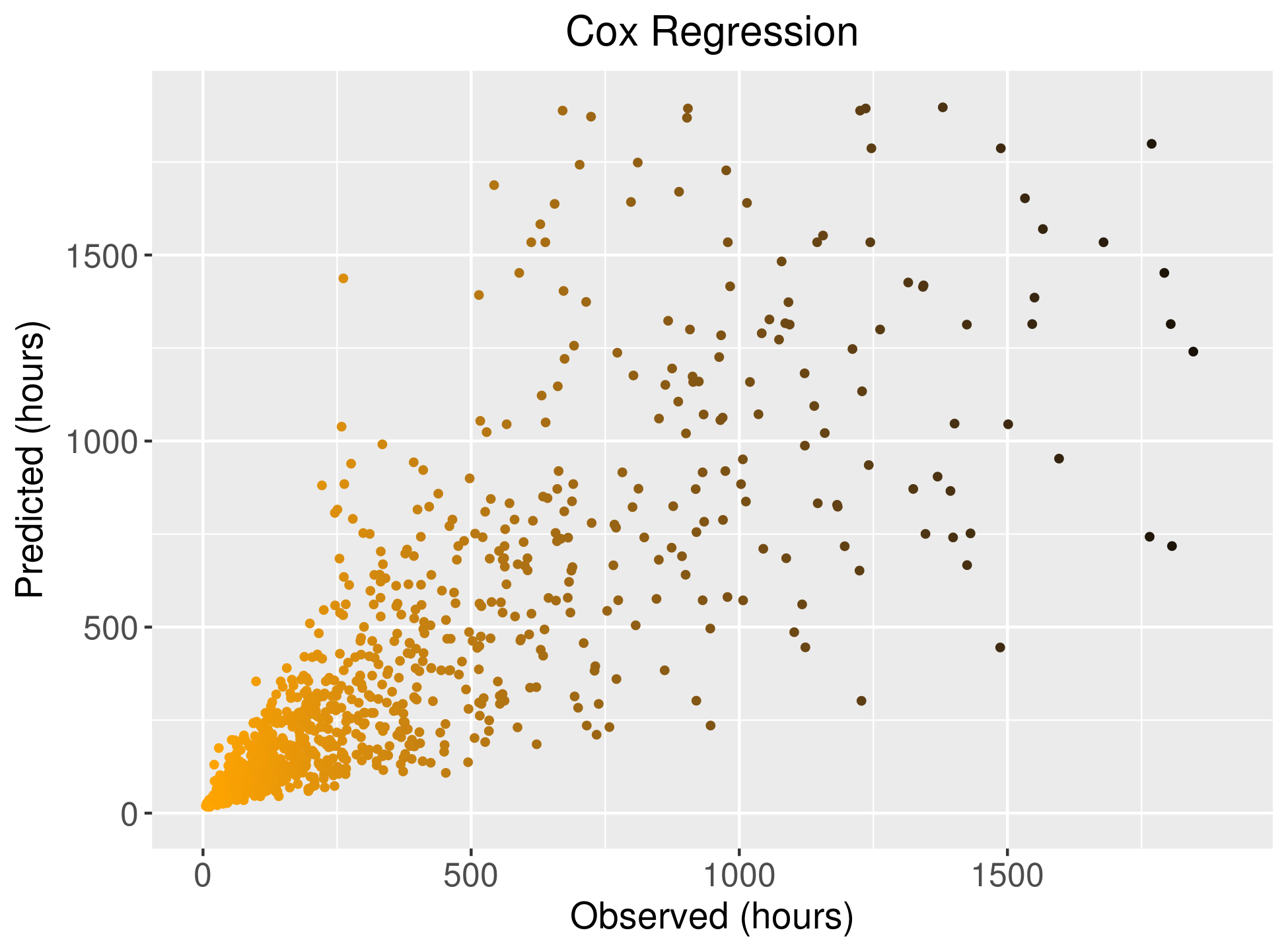}
   \includegraphics[width=0.49\textwidth]{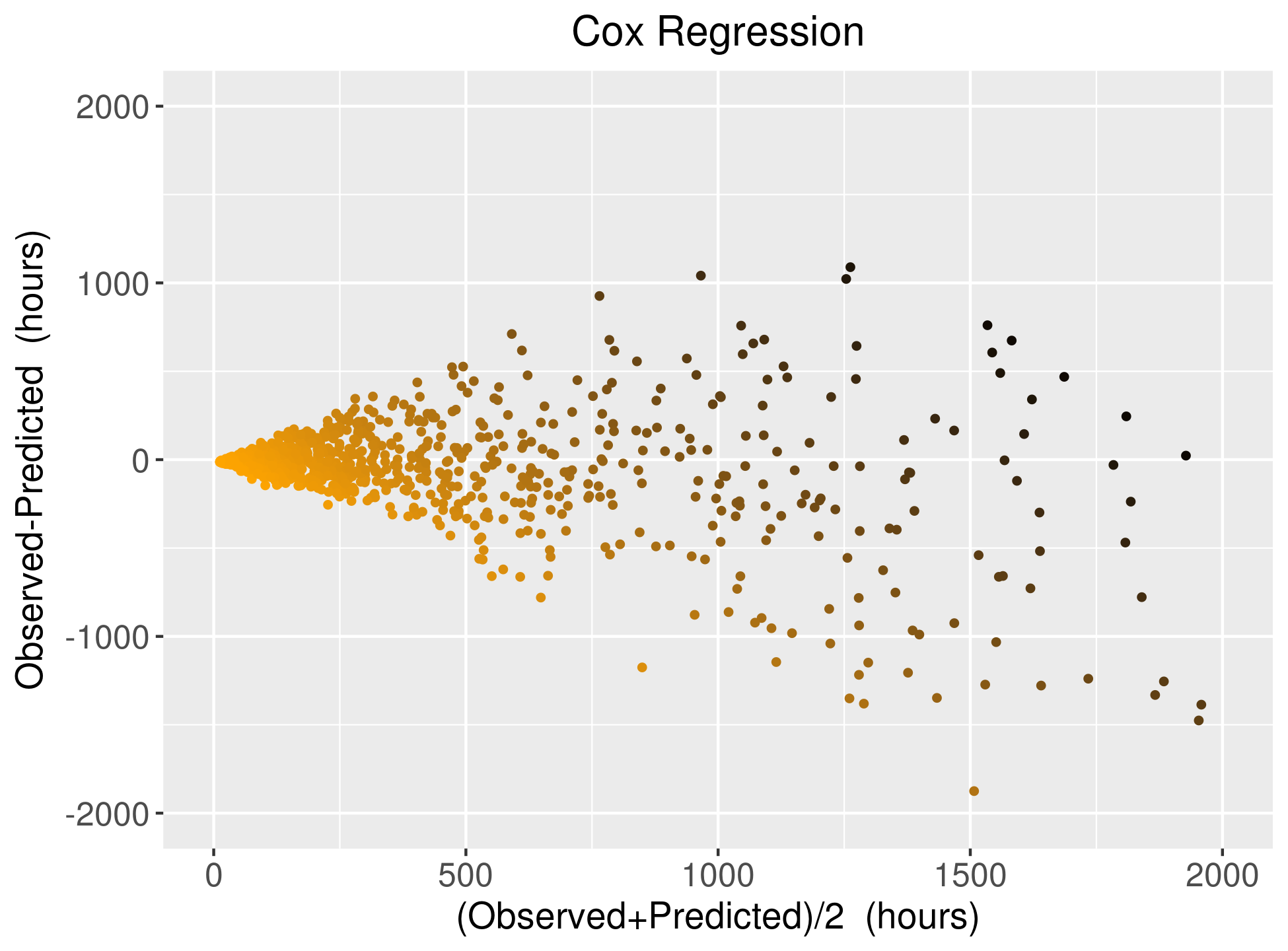} \\
\caption{Predicted median survival playtime vs.\ observed playtime (left) and 
relative deviation (right) for churned players, using the survival ensemble and 
Cox regression models.}
\label{playtime2}
\end{figure*}

\subsection{Dataset}
The data consisted of player action logs collected between 2014 and 2017 from a 
major mobile social game, \textit{Age of Ishtaria}, developed by Silicon Studio. The predictions were done on 
a subset of the most valuable players, who provide at least 50\% of the revenue (in this case 6.136 players). 

Since the model should be applicable to multiple types of games, we compute a 
set of input features that can easily be generalized to other games and properly 
captures the dynamics of the data. The feature calculation is parallelizable over 
all players, and the final dataset is small enough to be scalable to 
millions of players. 

In particular, from a player's action log, the daily logins, purchases,
playtime, and level-ups are extracted. These are commonly found in most
games and provide the essential information on playing behavior. For each of
these data sources, the mean is calculated over several different time
periods, namely over the player's first nine days, last nine days and full
lifetime. Further, the time elapsed until the first and last daily purchase is
calculated together with the amount spent on that day. Finally, to get the
current state of the player, the total purchases, total playtime, total logins
and current level are also added. This way of constructing the
feature set allows for an easy extension to include other data sources (e.g.\
click counts, experience gained, distance moved, etc.) and it is robust enough to describe the different variability of the data among games.

\subsection{Outcome}
Two additional models based on \cite{perianez2016churn} are implemented to 
perform predictions on the number of hours a user will play and the level at 
which they will quit. 
The models are trained on each of the following response variables:
\begin{itemize}
\item Playtime: How many seconds the user played the game. 
\item Level: Latest game level reached by the player.
\end{itemize}
In both cases, the censored variable is whether they churned or not (churn 
is defined as not having logged in for 9 days).

\subsection{Features}
For the predictors the most relevant variables from the dataset are selected for each of the models.
\begin{itemize}
\item Playtime model: Level, Days since last Purchase, First purchase amount, Last purchase amount, Purchases in the first 9 days, Loyalty index (number of days connected divided by the lifetime), Days since last level up. 
\item Level model: Lifetime, Days since last Purchase, 
First purchase amount, Last purchase amount, Purchases in the first 9 days, Loyalty index, Days since last level up.
\end{itemize}

\section{Results}

The model described above outputs a different survival probability curve for every player.
Figure \ref{level-users} illustrates the survival probability 
for three different users, each having a distinct survival expectation 
determined by their characteristics. In this case the first player is expected 
to churn at a very early level, while the third player would reach a much higher 
level. Similarly in the right figure, the survival prediction for three players is plotted in terms of playtime, distinguishing diffent degrees of playtime expectancy. 

The level and playtime prediction results are displayed in Figures~\ref{level} 
and \ref{playtime2}, respectively. We perform a comparison between the standard 
Cox regression \cite{Cox1972} and the survival ensemble model (an ensemble of 1.500 trees). When predicting 
at what level a player will leave the game, the Cox regression model has more 
difficulty capturing the temporal nonlinearity inherent to the problem (i.e. the time between levels is not uniform), as evidenced by a much larger spread in 
Figure~\ref{level}. The accuracy of the survival ensemble remains better 
throughout the entire level range, with points lying tightly along the identity 
line, i.e.\ with small differences between the predicted and observed values. 
The accuracy is reduced for higher levels, but this is explained by 
the fact that there is less data and the censorship increases (as there are 
fewer players at those levels). The same effect is observed for playtime in 
Figure~\ref{playtime2}: the predictions are most accurate for players with very 
little playtime, whereas the spread becomes more significant as playtime 
increases, which can be explained again by the fact that very few users have 
played so much.

\begin{figure}[ht!]
   \centering
   \includegraphics[width=0.49\textwidth]{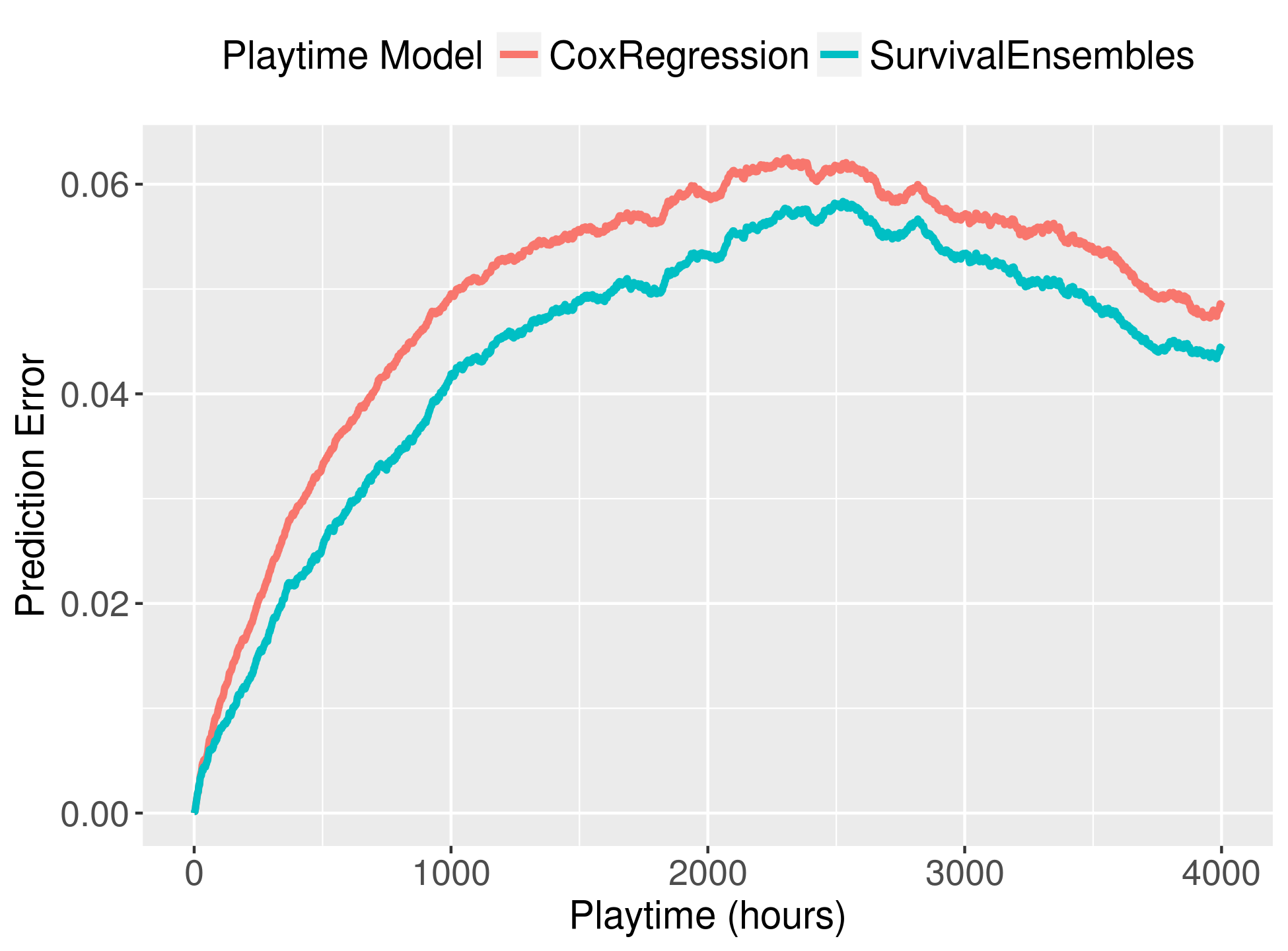}
   \includegraphics[width=0.49\textwidth]{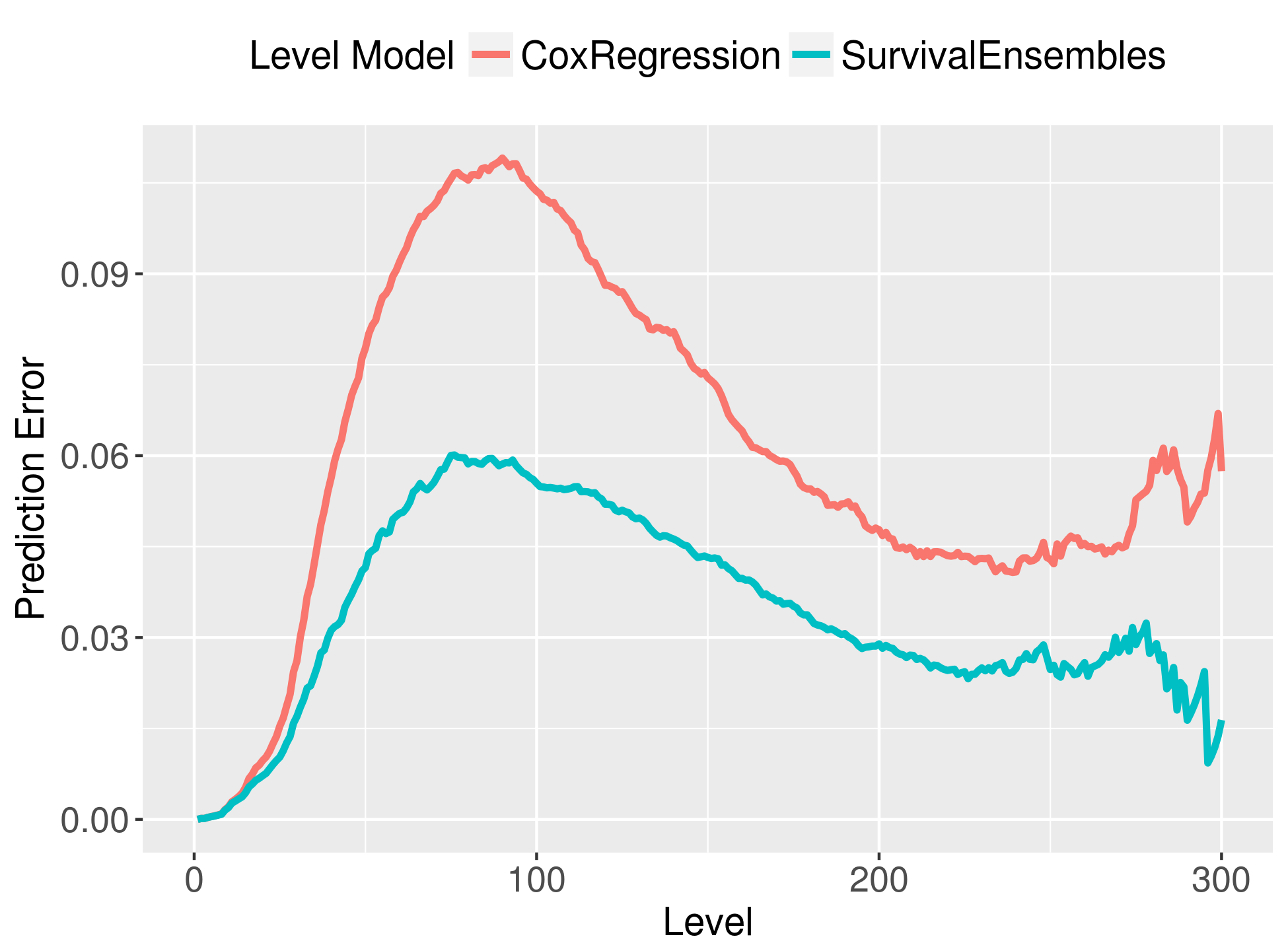}
 \caption{Playtime 
model (top) and level model (bottom) IBS error curves.}
 \label{errorMeasures}
\end{figure}

The integrated Brier scores (IBS)\cite{mogensen2012evaluating} calculated using 
bootstrap cross-validation with replacement with 1000 samples \cite{alfons2012cvtools} are listed 
in Table \ref{BrierTable_playtime}. It can be seen that the survival ensemble 
significantly outperforms the Cox regression model both for level and playtime 
prediction. Figure \ref{errorMeasures} also shows that the survival ensemble 
error is lower than that of Cox regression over the entire range of both 
playtime and levels. Figure \ref{errorMeasures} depicts the non-linearity of 
the time per level dimension (i.e. the time between levels is not equally distributed), which indeed will be diferent for every game.

\begin{table}
	\centering
    	\caption{Integrated Brier score (IBS) for level and playtime prediction}
	\begin{tabular}{ccc} \hline\noalign{\smallskip}
		Model             & IBS level & IBS playtime \\ \hline
		Survival Ensemble &  0.025 &  0.026\\  
		Cox regression    &  0.054 &  0.044\\  
		Kaplan-Meier      &  0.127 &  0.134\\ \hline 
\noalign{\smallskip}\hline 
	\end{tabular}
    	\label{BrierTable_playtime}
\end{table}

%

\section{Summary and Conclusions}
The results show that the method based on \textit{conditional inference survival 
ensembles} is able to model churn both in terms of playtime and level, 
predicting accurately at which level a player will leave and how long they will 
play. This indicates that the model is robust to different data distributions, 
and applicable to different types of response variables. While Cox regression 
did perform relatively well, it requires a lot of manual effort and also suffers 
from scalability issues, which makes it 
unsuitable for a production environment. On the other hand, the proposed survival
ensembles are easily adaptable to other type of games and uses a parallelized implementation that can be run not
only on multiple cores but also on multiple machines. This gives game developers 
the chance to efficiently obtain full survival probability curves for each 
player, and to predict in real time not only when a player will leave the game, 
but also at what level they will do it and how many hours they will play before 
quitting.   

%

\section*{Acknowledgements}
\label{con_8}
We thank our colleagues Sovannrith Lay and Peipei Chen for many useful 
discussions. We also thank Javier Grande for his careful review of the 
manuscript.

\bibliographystyle{abbrv}
\bibliography{bibliography}

\end{document}